\documentclass[letterpaper]{article} 
\usepackage{aaai25}  
\usepackage{times}  
\usepackage{helvet}  
\usepackage{courier}  
\usepackage[hyphens]{url}  
\usepackage{graphicx} 
\usepackage{natbib}  
\usepackage{caption} 
\usepackage{algorithm}
\usepackage{algorithmic}

\usepackage{booktabs}
\usepackage{multirow}
\usepackage{amsmath} 
\usepackage{subcaption} 

\usepackage[usenames,dvipsnames]{color}
\usepackage{colortbl}
\definecolor{mygray}{gray}{.9}
\definecolor{champagne}{rgb}{0.97, 0.91,0.81}
\definecolor{amber}{rgb}{1.0, 0.75, 0.0}
\definecolor{lightpink}{rgb}{1.0, 0.71, 0.76}
\usepackage{newfloat}
\usepackage{listings}
\DeclareCaptionStyle{ruled}{labelfont=normalfont,labelsep=colon,strut=off} 
\floatstyle{ruled}
\newfloat{listing}{tb}{lst}{}
\floatname{listing}{Listing}
\usepackage{color}
\newcommand{\yr}[1]{\textcolor{black}{#1}}
\newcommand{\yw}[1]{\textcolor{black}{#1}}

\title{SG-Splatting: Accelerating 3D Gaussian Splatting with Spherical Gaussians}
\author {
    Yiwen Wang\textsuperscript{\rm 1},
    Siyuan Chen\textsuperscript{\rm 1},
    Ran Yi\textsuperscript{\rm 1}
}
\affiliations {
    \textsuperscript{\rm 1}Shanghai Jiao Tong University\\
    wangyiwewn999@sjtu.edu.cn, gouliguojiashengsiyi@sjtu.edu.cn, ranyi@sjtu.edu.cn
}

\begin{document}

\maketitle

\begin{abstract}
3D Gaussian Splatting is emerging as a state-of-the-art technique in novel view synthesis, recognized for its impressive balance between visual quality, speed, and rendering efficiency. However, reliance on third-degree spherical harmonics for color representation introduces significant storage demands and computational overhead, resulting in a large memory footprint and slower rendering speed. We introduce SG-Splatting with Spherical Gaussians based color representation, a novel approach to enhance rendering speed and quality in novel view synthesis. Our method first represents view-dependent color using Spherical Gaussians, instead of three degree spherical harmonics, which largely reduces the number of parameters used for color representation, and significantly accelerates the rendering process. We then develop an efficient strategy for organizing multiple Spherical Gaussians, optimizing their arrangement to achieve a balanced and accurate scene representation. To further improve rendering quality, we propose a mixed representation that combines Spherical Gaussians with low-degree spherical harmonics, capturing both high- and low-frequency color information effectively. SG-Splatting also has plug-and-play capability, allowing it to be easily integrated into existing systems. This approach improves computational efficiency and overall visual fidelity, making it a practical solution for real-time applications.


\end{abstract}

\section{Introduction}
Novel view synthesis is \yr{an important} task in computer graphics and vision, focus\yr{ing} on generating new perspectives of a scene from a limited set of \yr{multi-view} images. This task addresses the challenge of creating realistic and immersive \yr{novel view images} without requiring extensive data collection or manual modeling. 
The advent of 3D Gaussian Splatting \cite{kerbl3Dgaussians} has injected new momentum into the field of novel view synthesis. 
Before \yr{the} introduction \yr{of 3D Gaussian Splatting}, NeRF \cite{mildenhall2020nerf} models, which rely on MLP for \yr{novel} view synthesis, were widely used but suffered from slow inference times and lengthy training processes. 
In contrast, 3D Gaussian Splatting leverages Gaussian splats to represent scene geometry and appearance, offering significant improvements in rendering speed, image quality, and training efficiency. This approach effectively addresses the limitations of NeRF, providing a more practical and scalable solution for applications requiring fast and high-quality view generation, such as virtual reality \cite{jiang2024vr-gs}, 3D\&4D content generation \cite{shao2023control4d, yang2023gs4d}, \yw{physical simulation \cite{xie2023physgaussian, zhong2024springgaus} and autonomous driving  \cite{Zhou_2024_CVPR}.}

\begin{figure}[t]
\centering
\includegraphics[width=0.4\textwidth]{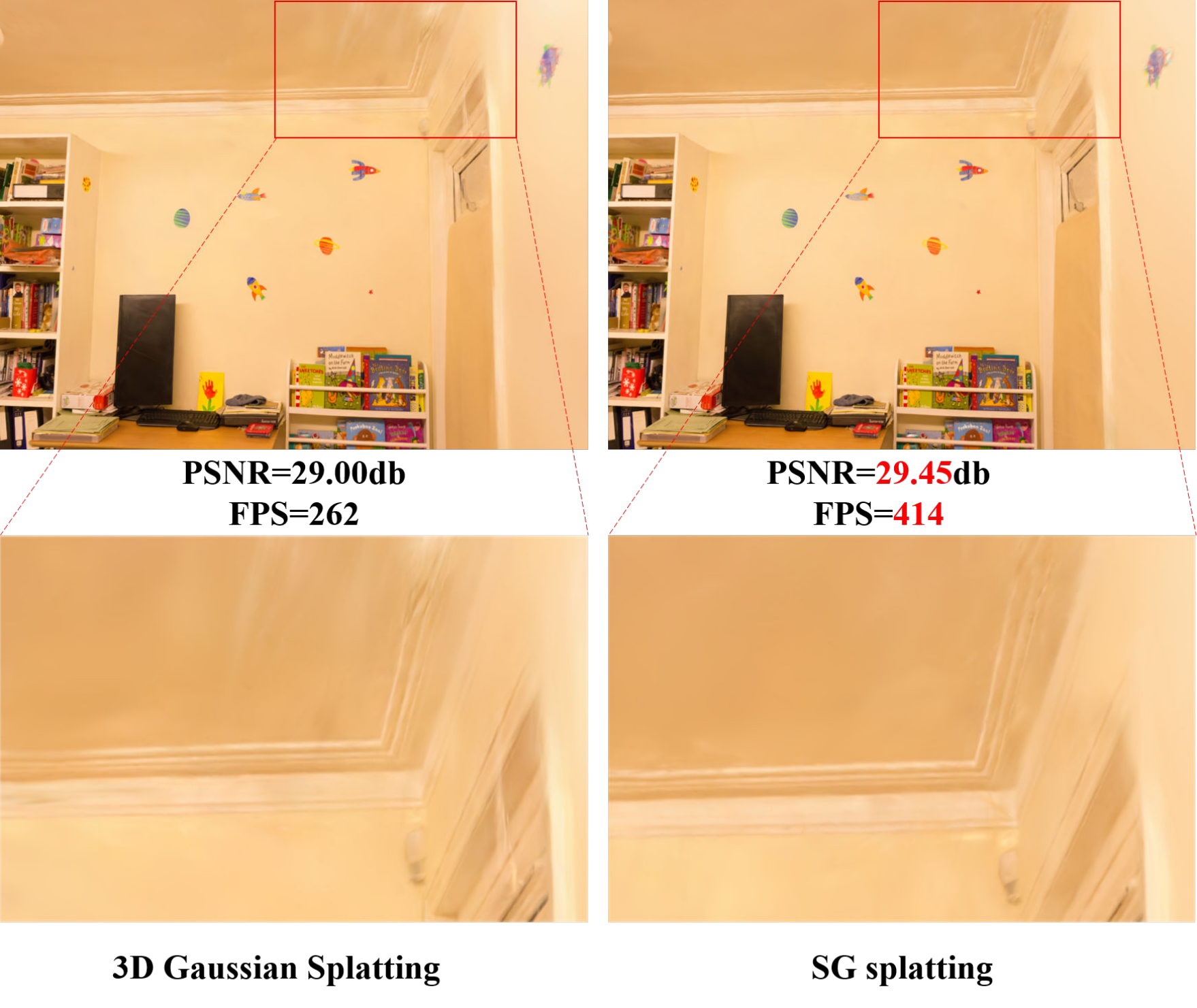}
\vspace{-0.15in}
\caption{\textbf{SG-Splatting: Spherical Gaussian-Based Splatting.} The top \yr{row} presents \yr{comparisons of SG-Splatting and 3D-GS on rendering results and} quantitative \yr{metrics on} the DeepBlending \cite{DeepBlending2018} dataset, demonstrating the ability to achieve rapid rendering. The bottom \yr{row} provides \yr{zoom-in} qualitative analysis of the rendered scenes, highlighting how 3D-GS exhibits noticeable artifacts, which are effectively minimized by our SG-Splatting approach. }

\label{fig:cover}
\end{figure}
Despite its advancements, 3D Gaussian Splatting faces significant challenges that limit its broader application, particularly in real-time rendering environments. One of the primary limitations is its reliance on third-\yr{degree} spherical harmonics \yr{(SH)} for color representation. While this approach allows for detailed and accurate color expression, it comes at the cost of a substantial memory footprint\yr{: each Gaussian requires $48$ SH coefficients, considering a scene with millions of Gaussians, the total number of SH coefficients is huge}. The high storage requirements not only \yr{consume a large amount of} computational resources, but also restrict the scalability of the technique in more complex or larger scenes. Additionally, the use of spherical harmonics introduces computational complexity that impacts the rendering speed\yr{}. The forward inference process requires considerable calculations, leading to slower rendering times, which can be a critical bottleneck in applications demanding real-time performance.

\yr{To address} the challenges posed by the large memory footprint and slow rendering speed of 3D Gaussian Splatting, several recent studies, including LightGaussian \cite{fan2023lightgaussian} and Compact3D \cite{navaneet2023compact3d}, have focused on compressing \yr{the model size of 3D Gaussian Splatting} to make it more efficient. These approaches primarily leverage traditional model compression techniques, such as knowledge distillation \cite{hinton2015distilling} and vector quantization \cite{jacob2018quantization}, to reduce the size of the model while attempting to maintain its visual quality. While these methods have shown promise in reducing the computational demands of 3D Gaussian Splatting, \yw{they overlooked the potential optimizations within the 3D Gaussian Splatting representation itself, which could lead to significant space compression and rendering acceleration.}

Upon further investigation, we identified that the most significant parameter overhead in 3D Gaussian Splatting stems from the coefficients of the spherical harmonics used for color representation. Specifically, the use of third-\yr{degree} spherical harmonics results in more than half of the model's parameters being allocated to storing these coefficients\footnote{\yr{In 3D-GS, each Gaussian has $59$ parameters for geometry and texture representation, where $48$ parameters are SH coefficients.}}, contributing to the substantial memory requirements and computational burden. 
To address this \yr{issue}, we propose \yr{\textbf{SG-Splatting}} 
\yr{with novel \textbf{Spherical Gaussians based color representation},}
which offers a more \yr{compact and} efficient representation. 
\yr{Firstly, we propose to represent view-dependent color using Spherical Gaussians, instead of three-degree SH, largely reducing the number of parameters required for color representation.}
\yr{Secondly}, we \yr{propose} the use of multiple Spherical Gaussians \yr{of orthogonal orientations,} to better simulate and optimize the scene, achieving a balance between reduced parameter overhead and enhanced rendering performance. 
\yr{Furthermore, we propose a mixed color representation that combines Spherical Gaussians with adaptive low-degree spherical harmonics, which combines the advantages of SG in high-frequency color representation, and the advantages of low-degree SH in low-frequency color representation, to further improve rendering quality.}
This approach 
accelerates the rendering process, making it more suitable for real-time applications.

Contributions \yr{of this paper} can be \yr{summarized as:} 
\begin{itemize}
    \item 
    \yr{Spherical Gaussians based Color Representation:}
    We propose \yr{SG-Splatting,} an innovative approach that 
    \yr{represents view-dependent color using Spherical Gaussians, instead of three-degree spherical harmonics,}
    significantly reducing the model's parameter overhead and improving computational efficiency.
    \item Effective Organization of Multiple Spherical Gaussians: To further enhance the model's performance, we develop a method for the effective organization of multiple spherical Gaussians, optimizing scene representation while maintaining high visual fidelity.
    \item \yr{Mixed Spherical Gaussians and low-degree spherical harmonics representation: We further propose a mixed color representation that combines the advantages of Spherical Gaussians in representing high-frequency color and the advantages of low-degree spherical harmonics in representing low-frequency color, to enhance rendering quality.}
    \item Plug-and-Play Capability: Our experiments show that SG-Splatting can be seamlessly integrated into other acceleration methods. This plug-and-play capability allows our approach to enhance existing techniques, improving 
    rendering speed while maintaining output quality.
\end{itemize}


\section{Related Works}
\subsection{Novel View Synthesis}
\yw{Novel view synthesis (NVS) traditionally relies on Structure-from-Motion (SfM) \cite{schoenberger2016sfm} and Multi-View Stereo (MVS) \cite{schoenberger2016mvs}, where SfM reconstructs a scene’s 3D structure from multiple images, providing the initial geometry for further processing.}

\yw{Neural Radiance Fields (NeRF) \cite{mildenhall2020nerf} brought a significant advancement to NVS by using a multi-layer perceptron (MLP) for volumetric scene modeling, though it suffers from slow training and rendering. Subsequent methods like Mip-NeRF 360 \cite{barron2022mipnerf360} improved image quality but at the cost of higher computational demands, while InstantNGP \cite{mueller2022instant} and PlenOctree \cite{yu2021plenoctrees} focused on speeding up training and rendering, respectively.}

\yw{Recently, 3D Gaussian Splatting (3D-GS) has emerged as a more efficient alternative for NVS, surpassing NeRF in both speed and quality. By approximating radiance fields with Gaussian components, 3D-GS enables real-time rendering, making it a transformative development for novel view synthesis.}


\subsection{3D-GS Fast Inference}

\yw{As an explicit rendering method, the inference speed of 3D Gaussian Splatting is closely related to the number of Gaussians used.} The inference process \yw{cost \(\epsilon\)} of 3D Gaussian Splatting can be approximately quantified as:
\begin{equation}
    \epsilon = N \times C\yr{,}
\end{equation}
where \( N \) represents the total number of Gaussians in the system, and \( C \) denotes the computational cost associated with processing each individual Gaussian. This relationship highlights the direct impact of the number of Gaussians on the overall computational expense, emphasizing the importance of efficient representation and processing within the 3D Gaussian Splatting framework.

\yr{To accelerate inference process}, \yw{GES \cite{Hamdi_2024_CVPR} reduces the number of Gaussians by optimizing Gaussian expressions using generalized exponential functions. LightGaussian \cite{fan2023lightgaussian} trims Gaussians based on their importance, determined by factors like opacity and volume. Compact3D \cite{navaneet2023compact3d} employs a learnable mask to prune Gaussians effectively. Mini-splatting \cite{fang2024mini} decreases sampling points by incorporating depth estimation, While Eagles \cite{girish2023eagles} and Compress3D \cite{niedermayr2023compressed} use quantized methods to compress attributes like color and direction of Gaussian primitives.} \yw{Although these methods have achieved notable success in speeding up the rendering process, they have not directly modified the most impactful component \yr{in Gaussian representation}, the spherical harmonics (SH). }

\subsection{Color Representation}

\yw{In novel view synthesis and computer graphics, color representation is crucial. Methods like NeRF use end-to-end MLPs to learn color information, but this approach demands significant computational resources, leading to long rendering times. To mitigate these challenges, SNeRG \cite{hedman2021snerg} separated appearance into diffuse and specular components, reducing reliance on large MLPs and cutting down on inference time. Building on this, PlenOctree \cite{yu2021plenoctrees} introduced spherical harmonics (SH) for color representation, effectively approximating spherical fitting and improving efficiency for scenes viewed from various angles.}

\yw{Other techniques, such as spherical functions \cite{allFreQuency,Xu13sigasia} and wavelet transforms \cite{allFreWavelet}, have also been explored for color representation and storage. These methods efficiently capture and compress color information while preserving detail, enabling faster rendering and reduced storage requirements. }

\yw{In comparison, Spherical Gaussians (SG) are the most flexible and computationally lightweight option. With fewer parameters and simpler computations, SGs are ideal for accelerating rendering while maintaining high quality, making them well-suited for novel view synthesis.}
\section{Methods}
\begin{figure*}[t]
\centering
\includegraphics[width=0.8\textwidth]{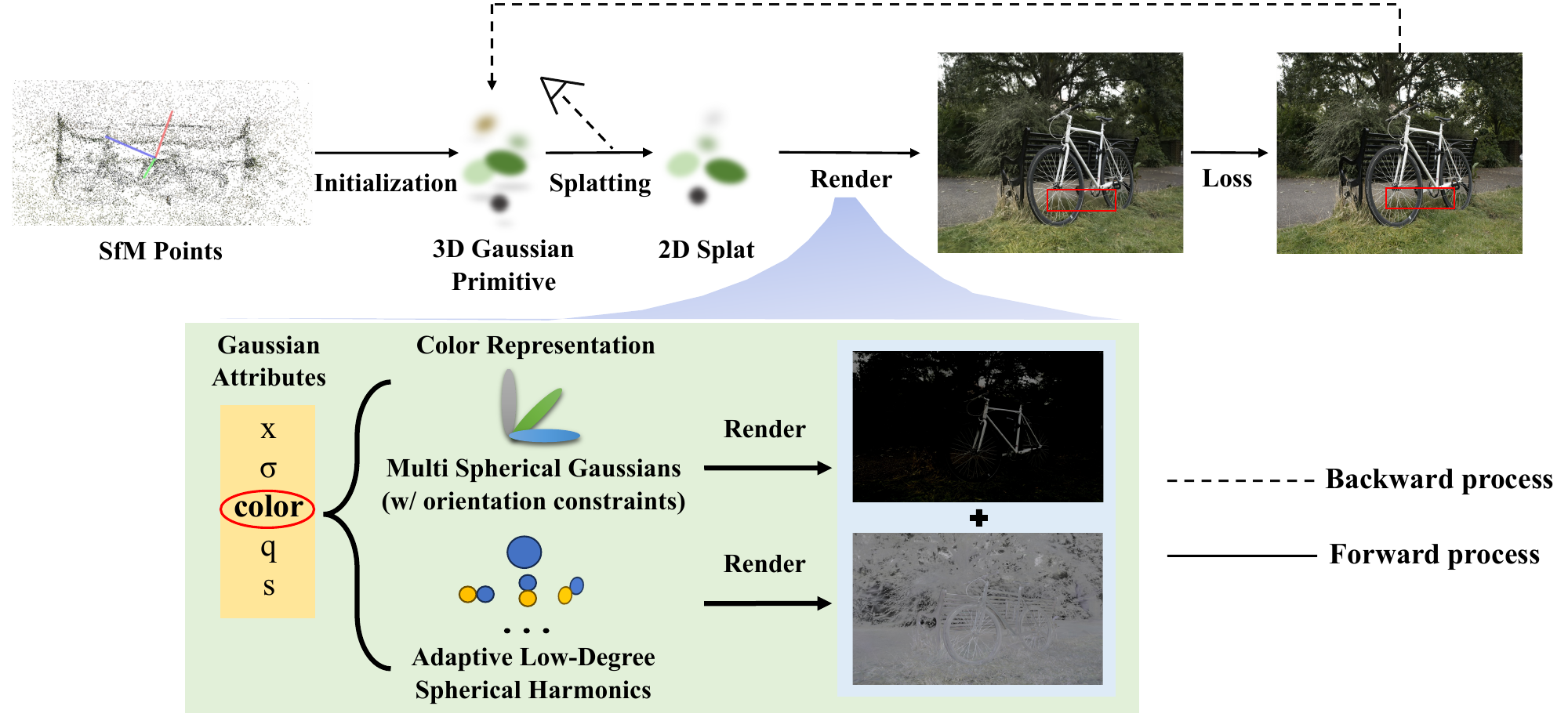} 
\caption{\yr{Our} SG-Splatting framework for novel view synthesis. The process begins with \yr{multi-view} images and corresponding camera views, which are used to initialize Gaussian primitives through \yr{SfM}. These primitives are characterized by their positions, covariance matrices, \yr{opacity} and color information. High-frequency color components are represented using Spherical Gaussians (SGs) to accurately capture view-dependent effects, while low-frequency color components are supplemented using \yr{diffuse or adaptive low-degree} Spherical Harmonics (SH) to provide a smooth and consistent representation. The final rendered output is compared with the ground truth to evaluate the \yr{method} performance.}

\label{fig:pipeline}
\end{figure*}

\subsection{Overview}
The original 3D Gaussian Splatting system, due to its reliance on three degree spherical harmonics, requires a large number of parameters to store the spherical harmonic coefficients. This results in a significant increase in the model's memory usage and also adversely affects the forward rendering speed. We propose SG-Splatting, a novel 3D Gaussian Splatting technique designed to overcome the limitations of traditional spherical harmonics \yr{by using} \textbf{Spherical Gaussians \yr{(SG) based color representation}}. 

As shown in Fig. \ref{fig:pipeline}, our SG-Splatting framework builds upon 3D-GS by 
\yr{incorporating Sperical Gaussians to form a more compact color representation.}
\yr{The inputs and output of our system are the same as original 3D-GS:} The input\yr{s} consists of \yr{multi-view} scene images and their corresponding camera parameters, \yr{and} the output is the final rendered image of the scene. The process begins by using the \yr{multi-view images} to generate an initial point cloud 
\yr{by SfM.}
This point cloud is then used to initialize elliptical Gaussians, which represent the scene's geometry and color\yr{, where the color adopts our SG-based representation}. These elliptical Gaussians are optimized by minimizing the loss between the rendered image and the ground truth, using alpha blending to produce the final scene renderings.

\yr{Different from original 3D-GS that uses three degree spherical harmonics (SH) for color representation,} to enhance rendering quality while maintaining efficiency, we \yr{further propose} a \textbf{mixed \yr{color representation}} that \textbf{combines Spherical Gaussians and \yr{adaptive} low-degree spherical harmonics}. 
\yr{1)} The \textit{\yr{high-frequency} color \yr{component}} is represented using Spherical Gaussians, \yr{which is more compact than SH, and excels at representing high-frequency color components, to} capture fine lighting and shading details. 
\yr{Using single SG requires only $7$ parameters to represent view-dependent color, much fewer than $45$ parameters required by three-degree SH. Furthermore, we propose to use multiple SGs of orthogonal orientations to complement each other and improve the modeling of color and light.}
\yw{
This orthogonal arrangement ensures that each Gaussian can effectively represent distinct aspects of the scene's color and light, minimizing redundancy and improving the overall rendering quality.} 
\yr{2)} Meanwhile, the \textit{\yr{low-frequency} color \yr{component}} is initially expressed using a simple 3-dimensional \yr{diffuse color (the zero degree SH component)}, with the option to replace this with \yr{adaptive} low-degree \yr{SH (0-2 degree)} for a better balance between rendering speed and quality. Although \yr{using low-degree SH alongside Spherical Gaussians} may slightly reduce rendering speed, it significantly improves the overall rendering quality, making this trade-off beneficial for scenarios where visual fidelity is a priority.



\subsection{Preliminaries}
3D Gaussian Splatting \yr{(3D-GS)} is an explicit method for modeling scenes in novel view synthesis 
\yr{using a set of 3D Gaussian ellipsoilds with geometric and texture properties.}


\yr{The geometry of each Gaussian is} parameterized by its \yw{center} position \yr{$X$} and a covariance matrix \yr{$\Sigma$}. The covariance matrix defines the shape, size, and orientation of the Gaussian, allowing it to model the spatial distribution of the data.\yw{Therefore, given a point $x$ in the 3D space, the Gaussian function is defined as:}
\begin{equation}
    G(x) = e^{-\frac{1}{2} {(x-\mathbf{X})} ^T\Sigma^{-1} (x-\mathbf{X})},
\end{equation}
\yr{where the covariance matrix is}
typically decomposed into \yr{a rotation matrix $R$ (represented by} a quaternion \yr{$q$}) and a \yr{scaling matrix $S$ (represented by} a scaling vector \yr{$s$}). 
This decomposition ensures that the Gaussian primitives are appropriately oriented and scaled in the 3D space\yr{:}
\begin{equation}
    \Sigma = RSS^TR^T.
\end{equation}

\yr{The texture information of each Gaussian consists of opacity $\sigma$ and spherical harmonics (to represent color).} For rendering, the 3D Gaussian primitives are first projected into 2D space from a given viewpoint \cite{zwicker2002ewa}. Once projected, the overlapping splats are accumulated and blended to create the final image.


\yr{\textbf{Spherical Harmonics (SH).}} 
To achieve high-quality rendering, particularly in capturing complex lighting and shading effects, 3D-GS utilizes three degree spherical harmonics \yr{(SH)} to represent the scene's color. 
\yw{Spherical harmonics have components that carry specific physical meanings, stemming from their mathematical origins in solving the angular part of Laplace's equation in spherical coordinates. These components are particularly useful in representing angular functions and capturing the directional properties of light and sound in various physical systems.} \yr{Specifically, 3D-GS uses the following formula to represent color by SH:}
\begin{equation}
c(\mathbf{v}) = \sum_{l=0}^{L} \sum_{m=-l}^{l} c_{lm} Y_{lm}(\mathbf{v}),
\end{equation}
\yw{where \( \mathbf{v} \) \yr{is} the input viewing direction, \( c(\mathbf{v}) \) is the color observed from that direction, \( c_{lm} \) \yr{is} the spherical harmonic coefficient, and \( Y_{lm}(\mathbf{v}) \) \yr{is} the spherical harmonic basis function \yr{of degree $l$ and order $m$}. The sum is taken over all degrees \( l \) and orders \( m \) up to a specified level \( L=3 \), capturing the complexity of the color variations. 
} 

Each \yr{SH} has a fixed and distinct shape. For a given spherical function, these \yr{SHs} need to be combined and their coefficients adjusted to accurately fit the function. This process involves fine-tuning each coefficient to capture the specific features and details of the spherical function being represented.
\yr{However, for three degree SH,} 
the entire 3D-GS system require\yr{s} $16$ \yr{SH} coefficients for each color channel\yr{, and $48$ SH coefficients in total}. This results in a significant increase in the model's parameter count, contributing to the substantial memory overhead and computational complexity associated with the system.

\subsection{Spherical Gaussian based Color Representation}
To address the inefficiencies associated with spherical harmonics, \yr{\textit{i.e.,} requiring $48$ SH coefficients for color representation,} we propose \yr{to} use Spherical Gaussians for color representation. Spherical Gaussians are more lightweight and flexible \yr{than} spherical harmonics, primarily because they require fewer parameters \yr{($10$ compared to $48$ of SH)} and can be easily adjusted to point in specific \yr{orientations}.  

\begin{figure}
\centering
\includegraphics[width=0.4\textwidth]{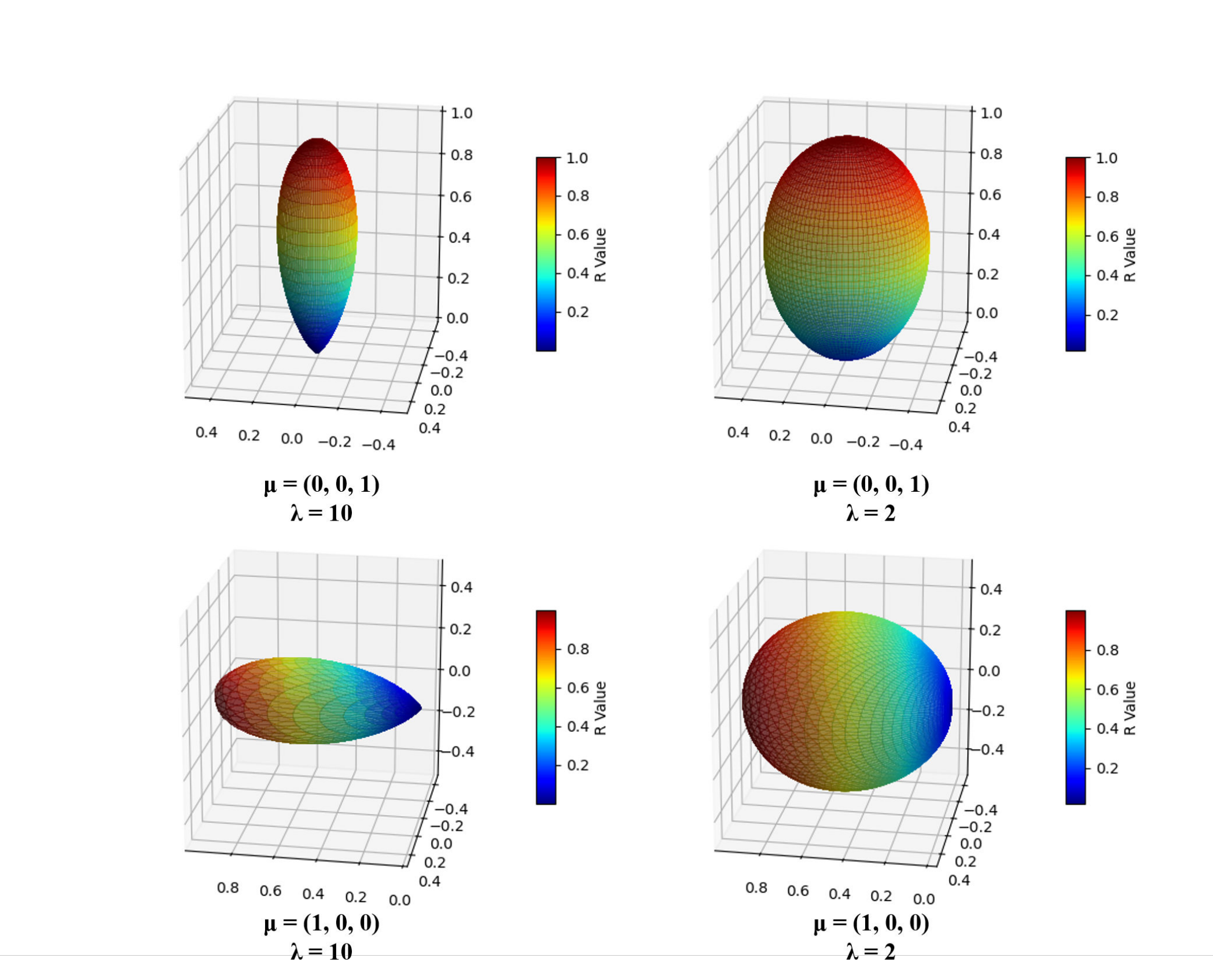}
\vspace{-0.15in}
\caption{\yw{Visualization of Spherical Gaussians \yr{(SG) and} how variations in \(\mu\) and \(\lambda\) affect the orientation \yr{and sharpness} of \yr{SG}. Comparing \yr{the top and bottom row} highlights the differences in \(\mu\), which changes the orientation of the \yr{SG}. Comparing from left to right illustrates the effect of \yr{sharpness} \(\lambda\); as \(\lambda\) increases, the \yr{SG} becomes more concentrated, resulting in a sharper peak.}}
\label{fig:spherical gaussian}
\end{figure}

\yr{\textbf{Background on Spherical Gaussians (SG).}}
\yw{As \yr{shown} in Figure \ref{fig:spherical gaussian}, 
the geometric shape of Spherical Gaussians \yr{resembles} lobes, \yr{allowing} for more precise control of light and color representation in different directions. 
\yr{Different from} spherical harmonics \yr{(SH) that} have fixed shapes and cannot be adjusted in form or position, Spherical Gaussians \yr{(SG)} offer flexibility \yr{in that: the orientation of lobe can be arbitrarily adjusted by paramter $\mu$, and the sharpness of lobe can be controlled by parameter $\lambda$ (see visualizations of different $\mu$ and $\lambda$ in Figure \ref{fig:spherical gaussian})}. 
This ability to dynamically adjust both orientation and sharpness gives Spherical Gaussians a distinct advantage in rendering applications, allowing for more accurate and versatile representations.}

\yr{\textbf{SG-based Color Representation.}}
\yr{Based on Spherical Gaussians,} the color $C$ of each Gaussian \yr{is} represented as:
\begin{equation}
C(d;{\alpha}, \lambda, {\mu}) = \alpha e^{\lambda (\yr{d}\cdot\mu -1 )}\yr{,}  
\end{equation}
where $d$ is the direction of the view, $\lambda$ controls the sharpness of the lobe, $\alpha$ represents the coefficients for the three color channels, and $\mu$ denotes the \yr{orientation} of the lobe's peak. A larger value of $\lambda$ results in a more concentrated lobe, increasing the sharpness of the effect.

\begin{figure*}
    
\centering
\includegraphics[width=0.8\textwidth]{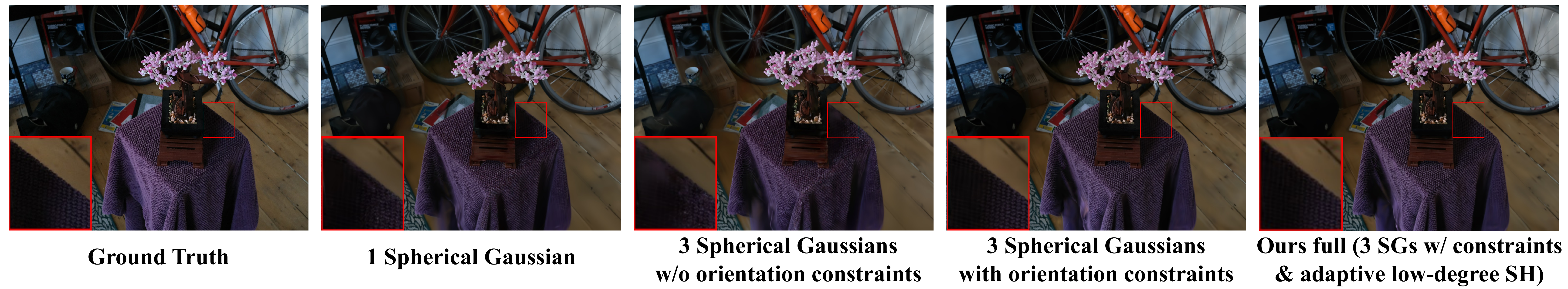}
\vspace{-0.1in}
\caption{\yw{Comparison of rendering results using different numbers of Spherical Gaussians (SGs). 
Using 3 SGs without orientation constraints can lead to inaccuracies \yr{(\textit{e.g.,}} incorrectly brightening shadowed areas), while with orthogonal orientation constraints, the rendering quality is improved.}
\yr{Ours full further integrates adaptive low-degree SH for better rendering quality.}}
\label{fig:3sg_suboptimal}
\end{figure*}

\yr{To further} enhance the accuracy of color representation, we decompose the color into two components: diffuse, and view-dependent. The diffuse component captures the inherent color of the surface that is independent of the viewing angle; while the view-dependent component captures effects \yr{such as} specular highlights, which \yr{vary depending} on the viewer's position relative to the light source. Therefore, the color \yr{is} represent\yr{ed} as:
\begin{equation}
C(d;{\alpha}, \lambda, {\mu}) = D +  \alpha e^{\lambda (\yr{d}\cdot\mu -1 )}\yr{,} 
\end{equation}
\yr{where} diffuse component $D$, and \yr{coefficients of each color channel} $\alpha$ are 3-dimensional vectors, \yr{orientation of Spherical Gaussian lobe's peak} $\mu$ is also 3-dimensional, and \yr{sharpness of Spherical Gaussian lobe} $\lambda$ is a scalar. 
\yr{Therefore, under SG-based representation,} each Gaussian requires only $10$ parameters \yr{($3$ for diffuse, $7$ for SG) to represent color}. 
This is significantly fewer than the number of parameters \yr{($48$)} needed for \yr{each} Gaussian \yr{under three degree SH-based color representation.} 
This reduction in parameter count highlights the efficiency of Spherical Gaussians, making them a much more compact and computationally efficient alternative\yr{, and allows for faster rendering and training speeds}.

\subsection{\yw{Multi-Axis Orthogonal Gaussians}}
\yw{While a single Spherical Gaussian significantly enhances rendering speed, it can face challenges in handling complex lighting scenarios, which may result in some limitations in rendering quality.} To address this \yr{issue}, we employ multiple Spherical Gaussians to improve the modeling of color and light, \yr{where multiple Spherical Gaussians have different orientations and complement each other}. 

However, simply accumulating multiple Spherical Gaussians led to suboptimal results (Fig. \ref{fig:3sg_suboptimal} \yw{third column}). The primary issue was that the Gaussians failed to distribute the scene's content effectively, resulting in clustering that further degraded the rendering quality. 
\yr{Spherical Gaussians can have arbitrary orientations of lobe (defined by $\mu$), which make them more flexible, but more}
difficult to \yr{control and} combine. 
\yr{\textit{E.g.,} in some scenes, without constraints on orientations of lobe, multiple Spherical Gaussians may be influenced by strong specular highlights in a certain orientation, and be optimized to similar orientations, failing to achieve the goal of complementarity.} 

\yr{\textbf{Multiple Orthogonal SGs.}}
\yw{To address this limitation, we introduce 
\yr{multiple orthogonal}
Spherical Gaussians, where the \yr{orientations} ($\mu$) \yr{of multiple gaussians} are orthogonal to each other. 
This orthogonal arrangement allows different Spherical Gaussians to complement one another more effectively, ensuring better alignment and coherence in the representation. 
As a result, this approach enhances rendering quality by maintaining a more structured and comprehensive depiction of the scene's color and light \yr{(Fig. \ref{fig:3sg_suboptimal} fourth column)}.} 

\yw{Each orthogonal direction acts as a basis vector. By using these \yr{orthogonal} directions, we achieve a more structured representation of the scene's color and light, reducing the likelihood of clustering and ultimately leading to improved rendering \yr{results}. 
\yr{Specifically}, the color $C$ \yr{is} represented as:}
\begin{equation}
C(d;{\alpha}, \lambda) = D + \sum_{i=1}^3 \alpha_i e^{\lambda_i (d\cdot\mu_i -1 )},   
\end{equation}
\yw{\begin{equation}\label{eq:orthogonal_basis}
\mu_i \cdot \mu_j = 
\begin{cases}
1, & \text{if } i = j,\\
0, & \text{if } i \neq j.
\end{cases}
\end{equation}}

\yw{\yr{Although} introducing multiple Spherical Gaussians increases the number of parameters, this increase is minimized by using orthogonal axes. 
Each Spherical Gaussian aligned along orthogonal axes 
requires $4$ parameters—\yr{coefficients for color} \(\alpha\) and \yr{sharpness} \(\lambda\). Additionally, 3 extra parameters are needed to define the orthogonal directions \yr{$\mu$}, resulting in a total of $15$ parameters for \yr{$3$} orthogonal \yr{SGs}. This is still far fewer than the $48$ parameters needed for third-degree spherical harmonics. This approach ensures that the \yr{multiple SGs} complement each other effectively, optimizing the overall representation while keeping the increase in complexity minimal. Moreover, our experiments show that the specific directions of the orthogonal axes are not crucial; the key is maintaining their orthogonality, which simplifies implementation (see \yr{Supplementary material} for experimental details).}

\subsection{Mixing Spherical Harmonics and Spherical Gaussians}
While the use of multiple Spherical Gaussians significantly accelerates the 3D Gaussian Splatting process, it can lead to some loss in rendering quality \yr{due to compressed color paramters}. To address this and further enhance the rendering quality, we incorporate \yr{adaptive} low-degree spherical harmonics \yr{(0-2 degree)} \yr{along with Spherical Gaussians for color representation}. 

\yw{\yr{Specifically, we adaptively adjust the degree used for SH based on the size of Gaussians.}
\yr{According to statistics on Gaussian radius (detailed in Supplementary material),}
we observed that most Gaussians in the scene are small. 
For small Gaussians, their coverage area is limited, and they do not exhibit significant view-dependent changes\yr{, mainly containing low-frequency color information}. 
As a result, there is no need to use high-degree spherical harmonics (SH) to fit them, and lower-degree SH is sufficient to capture their characteristics effectively.
} 
For these smaller Gaussians, we 
\yr{calculate} their size based on the matrix of their 2D projection, 
\yr{and} dynamically choose to use 0th, 1st, or 2nd degree spherical harmonics \yr{according to their sizes}. 
\yr{While for large Gaussians that are of a small proportion, we use 2nd degree spherical harmonics.} 
This adaptive approach allows us to \yr{achieve a} balance between rendering speed and quality, ensuring that each Gaussian is represented with the appropriate level of detail:
\begin{equation}
C(d;{\alpha}, \lambda) = SH +  \alpha e^{\lambda (\yr{d}\cdot\mu -1 )}. 
\end{equation}

\section{Experiments}

\begin{figure*}[t]
\centering
\includegraphics[width=0.88\textwidth]{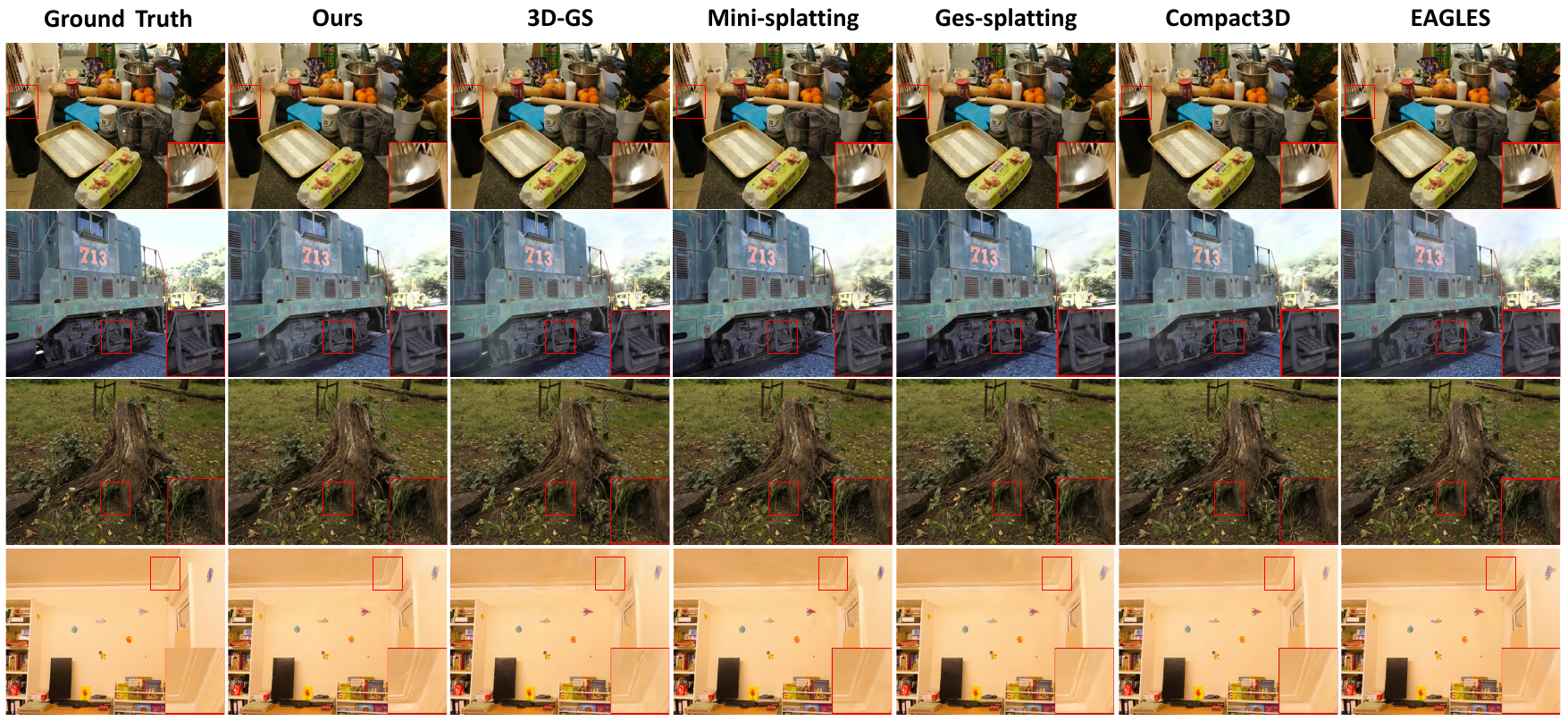} 
\caption{
\yr{Qualitative comparison of our SG-Splatting with baseline methods.}
}
\label{fig:qualitative_comparison}
\end{figure*}

\begin{table*}[t]
\renewcommand{\arraystretch}{1.2}
\resizebox{1.0\linewidth}{!}{
\begin{tabular}{c|ccccc|ccccc|ccccc}
\toprule
Dataset          & \multicolumn{5}{c|}{Mip-NeRF360} & \multicolumn{5}{c|}{ Tanks\&Temples} & \multicolumn{5}{c}{DeepBlending} \\
Matrices            & PSNR $\uparrow$  & SSIM $\uparrow$  & LPIPS $\downarrow$ & FPS $\uparrow$ & Train $\downarrow$   & PSNR $\uparrow$  & SSIM $\uparrow$  & LPIPS $\downarrow$ & FPS $\uparrow$ & Train $\downarrow$  & PSNR $\uparrow$   & SSIM $\uparrow$   & LPIPS $\downarrow$& FPS $\uparrow$& Train $\downarrow$    \\ \midrule
3D-GS             & \multicolumn{1}{>{\columncolor{amber}}c}{27.401}  & \multicolumn{1}{>{\columncolor{amber}}c}{0.814}  & \multicolumn{1}{>{\columncolor{amber}}c}{0.217}  & 247  & 25min29s  &\multicolumn{1}{>{\columncolor{amber}}c}{23.728} &	
                \multicolumn{1}{>{\columncolor{amber}}c}{0.845} &	\multicolumn{1}{>{\columncolor{amber}}c}{0.179} &	312 &14min40s  
                &29.508 &	0.900 & \multicolumn{1}{>{\columncolor{amber}}c}{0.247} &233 & 24min18s\\

Compact3D             & 26.969  & 0.797 & 0.245 & \multicolumn{1}{>{\columncolor{lightpink}}c}{293} & 35min41s &  23.323  & 0.831 & 0.202 & \multicolumn{1}{>{\columncolor{lightpink}}c}{383} & 20min41s & \multicolumn{1}{>{\columncolor{lightpink}}c}{29.749}  & \multicolumn{1}{>{\columncolor{lightpink}}c}{0.901} & 0.259 & \multicolumn{1}{>{\columncolor{lightpink}}c}{331} & \multicolumn{1}{>{\columncolor{lightpink}}c}{19min50s} \\ 
EAGLES          & 27.140   &0.805   &  0.239  & 155  & \multicolumn{1}{>{\columncolor{amber}}c}{21min55s} &  23.342   & 0.836  &   0.200 &  245 &  \multicolumn{1}{>{\columncolor{amber}}c}{11min22s} &  \multicolumn{1}{>{\columncolor{amber}}c}{29.951}   & \multicolumn{1}{>{\columncolor{amber}}c}{0.907}   & \multicolumn{1}{>{\columncolor{lightpink}}c}{0.249}  &  156 & \multicolumn{1}{>{\columncolor{amber}}c}{19min41s}  \\ 
SG-Splatting & \multicolumn{1}{>{\columncolor{lightpink}}c}{27.267}  &\multicolumn{1}{>{\columncolor{lightpink}}c}{0.813} & \multicolumn{1}{>{\columncolor{lightpink}}c}{0.218} & \multicolumn{1}{>{\columncolor{amber}}c}{334} &  \multicolumn{1}{>{\columncolor{lightpink}}c}{24min11s} & \multicolumn{1}{>{\columncolor{lightpink}}c}{23.465}  & \multicolumn{1}{>{\columncolor{lightpink}}c}{0.840} & \multicolumn{1}{>{\columncolor{lightpink}}c}{0.188} & \multicolumn{1}{>{\columncolor{amber}}c}{472} &  \multicolumn{1}{>{\columncolor{lightpink}}c}{13min21s}  &  29.570 & \multicolumn{1}{>{\columncolor{lightpink}}c}{0.901}  & \multicolumn{1}{>{\columncolor{amber}}c}{0.247} & \multicolumn{1}{>{\columncolor{amber}}c}{344} &  22min25s    \\
\midrule
Mini-splatting             & \multicolumn{1}{>{\columncolor{amber}}c}{27.345}  & 0.822 &\multicolumn{1}{>{\columncolor{amber}}c}{0.217} & 576 & 21min17s & \multicolumn{1}{>{\columncolor{amber}}c}{23.259} & \multicolumn{1}{>{\columncolor{amber}}c}{0.836}  & \multicolumn{1}{>{\columncolor{amber}}c}{0.202}   & 685 	  &  14min12s	 & \multicolumn{1}{>{\columncolor{amber}}c}{29.962}& \multicolumn{1}{>{\columncolor{amber}}c}{0.908}   &  \multicolumn{1}{>{\columncolor{amber}}c}{0.248} & 656   & 18min15s \\ 
SG-mini-Splatting &  26.933  & \multicolumn{1}{>{\columncolor{amber}}c}{0.835}  &0.222  & \multicolumn{1}{>{\columncolor{amber}}c}{681}  &    \multicolumn{1}{>{\columncolor{amber}}c}{21min05s} &  23.036  & 0.833   &  0.205 & \multicolumn{1}{>{\columncolor{amber}}c}{861} 	  &  	\multicolumn{1}{>{\columncolor{amber}}c}{12min37s}  &  29.916  & \multicolumn{1}{>{\columncolor{amber}}c}{0.908}   	 	& 	0.253 &  \multicolumn{1}{>{\columncolor{amber}}c}{773} &   \multicolumn{1}{>{\columncolor{amber}}c}{17min40s}    \\
\bottomrule
GES&  \multicolumn{1}{>{\columncolor{amber}}c}{26.901}  &  \multicolumn{1}{>{\columncolor{amber}}c}{0.794} &  \multicolumn{1}{>{\columncolor{amber}}c}{0.252} & 421 & 20min14s& \multicolumn{1}{>{\columncolor{amber}}c}{23.414}  & \multicolumn{1}{>{\columncolor{amber}}c}{0.837} & \multicolumn{1}{>{\columncolor{amber}}c}{0.198} & 518 & 11min02s&  29.628 & 0.901  & 0.252 & 398 & 19min04s \\ 
SG-GES  &  26.703 	 &0.791 	& 0.260 &  \multicolumn{1}{>{\columncolor{amber}}c}{520}   & \multicolumn{1}{>{\columncolor{amber}}c}{17min58s}   &  23.234  &   	0.832  & 0.202  &  \multicolumn{1}{>{\columncolor{amber}}c}{626}  & \multicolumn{1}{>{\columncolor{amber}}c}{10min01s}   &   \multicolumn{1}{>{\columncolor{amber}}c}{29.704} & \multicolumn{1}{>{\columncolor{amber}}c}{0.902}    & \multicolumn{1}{>{\columncolor{amber}}c}{0.251}  & \multicolumn{1}{>{\columncolor{amber}}c}{488}  & \multicolumn{1}{>{\columncolor{amber}}c}{17min37s}    \\
\bottomrule
\end{tabular}
}

\caption{Detailed comparison of \yr{our} SG-Splatting with 3D-GS and other baseline methods across various datasets. 
\yr{In comparison with the first group of baselines, we achieve the fastest rendering speed; and in the second group of baselines, after integrating our SG-Splatting into Mini-splatting and GES, the rendering speed is further improved, }
\yr{showing that our approach enhances efficieny while maintaining high visual quality.
The best and second best scrores in each group are highlighted in \colorbox{amber}{yellow} and \colorbox{lightpink}{pink}.}
}
\label{tab:overall comparison}
\end{table*}

\subsection{Experiment Settings}
\subsubsection{Datasets and Metrics.}
We conduct our experiments \yr{on} three \yr{commonly used} datasets: Mip\yr{-NeRF}360~\cite{barron2022mipnerf360}, Tanks\&Temples~\cite{Knapitsch2017}, and DeepBlending~\cite{DeepBlending2018}. 

To evaluate SG-Splatting, we use several metrics. Quality is measured using PSNR, SSIM \cite{SSIM}, and LPIPS \cite{zhang2018perceptual}\yr{, which} assess image fidelity, structural similarity, and perceptual quality, respectively. Rendering speed is measured in FPS, indicating the method's efficiency in real-time applications. We also \yr{report} training time 
to \yr{measure} computational efficiency.

\yw{All experiments are conducted on an NVIDIA RTX 3090 GPU to ensure fairness and avoid performance variations due to hardware differences.}

\subsubsection{Implementation.}
\label{experiments}
The learning rate hyperparameters for training are consistent with those used in 3D-GS to ensure 
\yr{fair comparison.}
The total number of training iterations is set to 30,000. After the initial 2,000 iterations, Spherical Gaussians are introduced into the training process with a learning rate of 0.0025.  This staged introduction helps in gradually refining the model’s ability to capture complex lighting and shading effects, leading to improved rendering quality. \yw{\yr{For multiple orthogonal SGs,} in our experiments, \yr{based on observations that the specific directions of orthogonal axes are not crucial while the key is the orthogonality (see experiments in Supplementary material),} we selected the orthogonal axes as \((1,0,0)\), \((0,1,0)\), and \((0,0,1)\) for ease of implementation.} 

\subsection{Comparisons}
\subsubsection{Baseline\yr{s}.}
\yw{We compare SG-Splatting with two groups of baseline methods. \yr{1)} The first group includes \yr{original 3D-GS,} Compact3D \cite{navaneet2023compact3d} and Eagles \yr{\cite{girish2023eagles}}, which primarily focus on post-processing trained models through \yr{model} compression techniques to reduce the number of Gaussian primitives and improve efficiency.} \yr{2)} \yw{The second group, including 
Mini-splatting \cite{fang2024mini} and GES \cite{Hamdi_2024_CVPR}, optimizes the 3D-GS model and training process to enhance rendering quality and speed. Our SG-Splatting can be seamlessly integrated into these approaches, offering additional improvements in efficiency. 
}
\subsection{Results and Analysis}
Table \ref{tab:overall comparison} presents a comprehensive comparison, \yr{including} the performance of our SG-Splatting, \yr{and} the effects of integrating SG-Splatting into other existing approaches. \yr{Fig.~\ref{fig:qualitative_comparison} shows the qualitative comparison results.}
\yw{In \yr{comparison with} the first \yr{group} of \yr{baselines}, our method \yr{achieves the fastest rendering speed, $1.4\sim 1.5$ times faster than original 3D-GS; with} rendering quality slightly lower than the original 3D-GS, but consistently outperform\yr{ing} other compression-based post-processing methods, particularly in the most complex scenes from Mip-NeRF360.}
\yw{In the second \yr{group} of \yr{baselines}, after \yr{integrating} our \yr{SG-based color representation into these methods (
Mini-splatting and GES)}, we observe a significant improvement in rendering speed across \yw{most} scenes, with only a minor reduction in rendering quality. This demonstrates the effectiveness of our approach in improving efficiency while maintaining high visual fidelity.}

Furthermore, SG-Splatting demonstrates a storage size reduction of approximately $46.7\%\sim 47.3\%$ between datasets compared to 3D Gaussian Splatting, further enhancing its practicality for memory-constrained environments.

\begin{table}[t]
\renewcommand{\arraystretch}{1.2}
\resizebox{1.0\linewidth}{!}
{
\begin{tabular}{c|ccccc}
\toprule
Ablation Setup          & PSNR $\uparrow$	 & SSIM $\uparrow$	& LPIPS $\downarrow$ & FPS $\uparrow$ & Size(MB) $\downarrow$\\
\toprule
3D Gaussian Splatting& 27.410 & 0.814 &0.217 & 247  & 781.46\\
\midrule
\yw{3DGS $+$ Multi-Axis Orthogonal SG} & {26.845}  & 0.803  &0.229   & 371 & 364.01 \\
3DGS + Low Degree SH& 27.206 & 0.812 & 0.219 &341  & 517.45\\
\yw{3DGS $+$ Multi-Axis Orthogonal SG $+$ Mix SH (Ours)} 
& \yr{27.267}  & \yr{0.813}  & \yr{0.218}   & 334  & 416.02\\
\bottomrule
\end{tabular}
}
\caption{Ablation studies on the Mip-NeRF360  dataset.}
\label{tab:ablation}
\end{table}

\begin{table}[t]
\renewcommand{\arraystretch}{1.2}
\resizebox{1.0\linewidth}{!}{
\begin{tabular}{c|ccc}
\toprule
Dataset        & Mip-NeRF360 	 &Tanks\&Temples  & DeepBlending 	\\
\toprule
3D Gaussian Splatting & 781.46 &434.10 & 662.26\\
SG-Splatting & 416.02 & 228.78  & 357.45 \\
\bottomrule
\end{tabular}
}
\caption{Comparison of storage size (in MB) for 3D Gaussian Splatting and SG-Splatting. }
\label{tab:size_ana}
\end{table}

\subsection{Ablation Studies}
We conducted ablation studies by progressively adding our components to the base 3D-GS model to evaluate their impact (Table~\ref{tab:ablation} and Fig.~\ref{fig:3sg_suboptimal}). Specifically, we tested the effects on rendering speed, rendering quality, and model size, evaluating how each component contributes to overall performance.

\textit{3DGS + Multi-Axis Orthogonal SG:} Introducing orthogonal SGs achieves a significant improvement in rendering speed and model size, at the cost of a slight decrease in rendering quality. Specifically, the rendering speed increases from 247 to 371 FPS, while the model size is reduced from 781.46 to 364.01 MB. Although there is a minor drop in PSNR (from 27.410 to 26.845) and SSIM (from 0.814 to 0.803), the trade-off results in a much more efficient model. The orthogonal arrangement of the SGs still captures the details of the scene effectively, enabling faster rendering and better storage efficiency.

\textit{3DGS + Low Degree SH:} To rigorously evaluate the effectiveness of spherical Gaussians (SG), we also conducted experiments using low-degree spherical harmonics (SH). This setup simplifies the representation, achieving a moderate improvement in rendering speed (341 FPS compared to the original 247 FPS) and a reduction in model size to 517.45 MB. 

\textit{3DGS + Multi-Axis Orthogonal SG + Mix SH (Ours):} In this configuration, we combine the advantages of low-degree spherical harmonics (SH) and spherical Gaussians (SG) to achieve a balanced trade-off between rendering quality, speed, and model size. Compared to the original 3D-GS, our method maintains comparable rendering quality (PSNR: 27.267 vs. 27.410, SSIM: 0.813 vs. 0.814) while significantly improving efficiency. Specifically, our method achieves a higher rendering speed (334 FPS compared to 247 FPS) and a notably smaller model size (416.02 MB compared to 781.46 MB). Using the complementary strengths of SG and low-degree SH, this approach provides an effective solution for scenarios requiring both high-performance and compact model storage.

\section{\yr{Conclusion}}
In this paper, we \yr{propose} SG-Splatting, a 3D Gaussian Splatting technique designed to address the limitations of 3D Gaussian Splatting, \yr{especially} the excessive number of third-\yr{degree} spherical harmonic coefficients that hinder\yr{s} rendering speed. Our approach utilizes Spherical Gaussians \yr{for a more compact color representation,} to mitigate the issues of coefficient overload and slow inference times. Additionally, we incorporate multiple Spherical Gaussians and mix them with \yr{adaptive} low-\yr{degree} spherical harmonics to enhance rendering quality while maintaining fast rendering speeds. Moreover, SG-Splatting is designed to be plug-and-play, allowing it to be easily integrated into other acceleration techniques without compromising performance.

\bibliography{aaai25}
\section*{Reproducibility Checklist}

\begin{enumerate}
    \item This paper:

\begin{itemize}
    \item Includes a conceptual outline and/or pseudocode description of AI methods introduced (yes)
    \item Clearly delineates statements that are opinions, hypothesis, and speculation from objective facts and results (yes)
    \item Provides well marked pedagogical references for less-familiare readers to gain background necessary to replicate the paper (yes)
\end{itemize}

\item Does this paper make theoretical contributions? (yes)

If yes, please complete the list below.
\begin{itemize}
    \item All assumptions and restrictions are stated clearly and formally. (yes)
    \item All novel claims are stated formally. (yes)
    \item Proofs of all novel claims are included. (yes)
    \item Proof sketches or intuitions are given for complex and/or novel results. (yes)
    \item Appropriate citations to theoretical tools used are given. (yes)
    \item All theoretical claims are demonstrated empirically to hold. (yes)
    \item All experimental code used to eliminate or disprove claims is included. (yes)
\end{itemize}

\item Does this paper rely on one or more datasets? (yes)

If yes, please complete the list below.
\begin{itemize}
    \item A motivation is given for why the experiments are conducted on the selected datasets. (yes)
    \item All novel datasets introduced in this paper are included in a data appendix. (NA)
    \item All novel datasets introduced in this paper will be made publicly available upon publication of the paper with a license that allows free usage for research purposes. (NA)
    \item All datasets drawn from the existing literature (potentially including authors’ own previously published work) are accompanied by appropriate citations. (yes)
    \item All datasets drawn from the existing literature (potentially including authors’ own previously published work) are publicly available. (yes)
    \item All datasets that are not publicly available are described in detail, with explanation why publicly available alternatives are not scientifically satisficing. (NA)
\end{itemize}

\item Does this paper include computational experiments? (yes)

If yes, please complete the list below.
\begin{itemize}
    \item Any code required for pre-processing data is included in the appendix. (yes)
    \item All source code required for conducting and analyzing the experiments is included in a code appendix. (yes)
    \item All source code required for conducting and analyzing the experiments will be made publicly available upon publication of the paper with a license that allows free usage for research purposes. (yes)
    \item All source code implementing new methods have comments detailing the implementation, with references to the paper where each step comes from. (yes)
    \item If an algorithm depends on randomness, then the method used for setting seeds is described in a way sufficient to allow replication of results. (NA)
    \item This paper specifies the computing infrastructure used for running experiments (hardware and software), including GPU/CPU models; amount of memory; operating system; names and versions of relevant software libraries and frameworks. (partial)
    \item This paper states the number of algorithm runs used to compute each reported result. (yes)
    \item Analysis of experiments goes beyond single-dimensional summaries of performance (e.g., average; median) to include measures of variation, confidence, or other distributional information. (no)
    \item The significance of any improvement or decrease in performance is judged using appropriate statistical tests (e.g., Wilcoxon signed-rank). (no)
    \item This paper lists all final (hyper-)parameters used for each model/algorithm in the paper’s experiments. (yes)
    \item This paper states the number and range of values tried per (hyper-) parameter during development of the paper, along with the criterion used for selecting the final parameter setting. (partial)
\end{itemize}

\end{enumerate}

\end{document}